\documentclass[manuscript,nonacm]{acmart}

\setcopyright{acmlicensed}
\copyrightyear{2026}
\acmYear{2026}
\acmDOI{XXXXXXX.XXXXXXX}
\acmISBN{978-1-4503-XXXX-X/2018/06}

\begin{document}

\title{Invisible Strings: Deriving Puppetry Principles and their Hidden Connections to Robot Behavior Design}

\author{Claire Lewis}
\email{claire_lewis@mines.edu}
\author{Sawyer Collins}
\email{sawyer.collins@mines.edu}
\author{Alyssa Hanson}
\email{abhanson@mines.edu}
\affiliation{%
  \institution{Colorado School of Mines}
  \city{Golden}
  \state{Colorado}
  \country{USA}
}

\author{Johanna Smith}
\email{johanna@csusb.edu}
\affiliation{%
  \institution{California Sate University San Bernadino}
  \city{San Bernadino}
  \state{California}
  \country{USA}
}

\author{Tom Williams}
\email{twilliams@mines.edu}
\affiliation{%
  \institution{Colorado School of Mines}
  \city{Golden}
  \state{Colorado}
  \country{USA}
}

\renewcommand{\shortauthors}{Lewis et al.}

\begin{abstract}
When designing robots' nonverbal behaviors, many researchers have turned to arts-based insights, such as Disney's Animation Principles. 
Yet, while these principles bear key insights into the design of like-life characters, their application to robot design is inherently limited, in part because animation is not constrained by real-world physics, and in part because animation principles focus on low level animation mechanics and not high-level design considerations for physically embodied, interactive characters. In contrast, little attention has been paid to art forms like puppetry, despite their long history of exploring morphological, behavior, and interaction design of physically embodied, interactive characters. 
As such, in this work we leverage puppetry texts and practicing puppeteers' expert knowledge knowledge
to derive a set of puppetry principles with key insights for robot design. As we show, these insights go beyond --- and uniquely complement --- the prior insights provided by theater, dance, and animation. 
\end{abstract}




\maketitle

\section{Introduction}
Robot designers regularly leverage nonverbal behaviors like gaze and gesture to enable more clear and lifelike communication. For example, designers often use gesture and body language to direct people's attention, communicate urgency, and increase likability~\cite{speedgesture, positivegesture, brooks2006working}. 
Similarly, designers use gaze  to direct attention, and to enhance team performance~\cite{ito2004gaze, admoni2017gaze, admoni2013gaze}. 
Finally, designers have recently explored the use of breath to signal robots' animacy and to communicate emotions implicitly~\cite{klausen2022}. All of these nonverbal cues, if effectively designed, can work together to create a lifelike embodied character. 

To design effective nonverbal cues, roboticists have often turned to art forms like theater, dance, and animation, whose practitioners and theorists have developed key theories for encoding and designing nonverbal behaviors in non-robotic contexts.
For example, roboticists have leveraged theatrical knowledge on the design and performance of interactions. However, theater-based design rarely focuses on designing specific motions. 
Similarly, roboticists have leveraged dance knowledge on the design and performance of specific movements~\cite{bacula2024dancing}. However, dance-based theories are often focused specifically on anthropomorphic actors. 
Finally, roboticists have leveraged Disney's 12 Animation Principles~\cite{Disney} to inform the animation of low-level robotic motions~\cite{ribeiro2012illusion,yunus2020cues, ribeiro2012illusion}. However, animation-based insights often leverage the fact that animations are not  constrained by the laws of physics.
Accordingly, we argue that while HRI researchers have successfully leveraged key arts-based insights in prior work, the nature of the specific art forms leveraged has led to gaps in their applicability. 

In contrast, in this work we consider the often overlooked art form of puppetry, and consider how it might fill the specific gaps left by these other art forms~\cite{JohannaSmith}.
Puppetry is an art form that bears striking similarity to robotics. 
Like robotics, puppetry centers the design of specific motions for not-fully anthropomorphic embodied artifacts that are subject to the laws of physics. Moreover, like roboticists, puppeteers must reason carefully about not just behavior design, but also the morphological design necessary to generate specific behaviors, and the interaction design between both puppet and audience --- and between puppet and puppeteer.
As such, we argue that puppetry insights might yield key insights for robot behavior design

Yet unlike fields like theatre, dance, and animation, few comprehensive sets of puppetry principles exist for ready leverage in robotics. The most formalized set of examples may be the Handspring Puppet Company's eleven principles for puppetry~\cite{handspring}. Yet even those principles focus primarily on the puppeteer rather than the puppet itself, and (as we will show) are noncomprehensive, lacking a number of widely accepted puppetry practices. 
Accordingly, in this work we derive a set of 8 novel puppetry principles derived from a range of puppetry texts, and supplemented by expert puppeteering knowledge. 
As we will show, the insights encoded in these 8 puppetry principles go
beyond --- and uniquely complement --- the prior insights provided
by theater, dance and animation.

\section{Background}
\subsection{Non-verbal Communication in Robotics}

Human-human interaction involves explicit and implicit communication~\cite{implicitHuman, goals}. Explicit communication involves the use of intentionally communicative behaviors like language, deictic gaze and gesture. Implicit communication involves the types of behaviors that humans typically execute subconsciously, such as beat gestures, non-deictic gaze,  body language, and breath~\cite{implicitExplicit, languageExIm}. To enable effective human-robot interactions, robots must effectively wield both explicit and implicit communication~\cite{implicitExplicit}, as well as communicative behaviors that fall along the spectrum between explicit and implicit~\cite{lessObnoxious}.
Across both explicit and implicit dimensions of communication, nonverbal behaviors like body language and gesture, gaze, and breath are critical in robot design.

\subsubsection{Body Language and Gesture}
Physical movements like gesture are critical for both human and robot communication. 
Deictic gestures (both physical and virtual~\cite{han2023crossing,brown2023best,williams2019mixed}) in particular are highly effective at shifting interlocutor' attention~\cite{positivegesture,scheutz2007first,brooks2006working,sauppe2014robot} while building rapport~\cite{breazeal2005effects} and increasing likability and warmth~\cite{brown2023best}. 
Meanwhile, a robot's body language can tell humans when it is unsafe to enter a certain proximity to the robot~\cite{collabgesture}.
Moreover, robots' physical movements can shape human behavior. For example, the faster the pace of a robot's gesture or body language, the quicker humans respond to the robot, perhaps due to the sense of urgency the robot's movement creates~\cite{speedgesture}. 

\subsubsection{Gaze}
Like gesture, robot gaze can also be used to signal and direct attention, and to communicate emotional state, engagement, and attentiveness~\cite{admoni2017gaze,admoni2013gaze,tapus2007gaze}.
Mutual gaze (direct eye contact) in particular has been found to increase perceived social awareness and intentionality of robots~\cite{ito2004gaze}, as well as engagement and task performance~\cite{fischer2018gaze}. Similarly, mutual gaze can help capture human attention, leading to human-robot conversation initiation~\cite{satake2009gaze}. Even for robots without facial features, simple movements like head turns can enhance human engagement~\cite{bruce2002gaze}. Gaze behaviors can even enhance interaction fluidity, such as the timing and coordination of object handovers~\cite{moon2014gaze}.
Mutual gaze (direct eye contact) in particular has been found to increase perceived social awareness and intentionality of robots~\cite{ito2004gaze}, as well as engagement and task performance~\cite{fischer2018gaze}. Similarly, mutual gaze can help capture human attention and encourage humans to initiate conversation~\cite{satake2009gaze}. Even for robots without facial features, simple movements like head turns can significantly enhance human engagement~\cite{bruce2002gaze}. 

\subsubsection{Breath}
Breath is a powerful implicit signifier of animacy, but is also used as part of explicit communication, e.g., when forming utterances~\cite{nakatani2018}. Researchers have leveraged this by incorporating utterance-based breathing patterns into social robots~\cite{nakatani2018,klausen2022}, and found that the rate of a robot’s breathing significantly influences human perceptions of its emotional state, including perceived levels of pleasure and arousal~\cite{klausen2022}. Breath has also been explored as a tool for reducing users' anxiety~\cite{asadi2022breath} and for enhancing users' athletic performance~\cite{sawchuk2024breath}. 
Recent work has shown how participants co-design robot breathing patterns to encourage "focusing", "grounding", "anchoring", and "tethering"~\cite{matheus2022breath}. 

Given the importance of body language, gestures, gaze, and breath to robot design, much human-robot interaction research has focused on the design of these these nonverbal behaviors. However, due to the inherently embodied nature of these behaviors, nonverbal behavior design is quite challenging. While research on human-robot dialogue has been able to leverage insights from the linguistics literature, research on robot nonverbal behaviors more commonly comes from the arts, especially areas like theater, dance, and animation, where artists explicitly focus on understanding and intentionally conveying the movement and physical behaviors of human (and non-human) characters.

\subsection{Robot Design Insights from the Arts}

\subsubsection{Theater}
Because theater seeks to convey convincing human-human interactions, roboticists are increasingly drawing on methods from theater when designing human-robot interactions~\cite{schleicher2010bodystorming,abtahi2021presenting,porfirio2019bodystorming,alves2021children,pelikan2023designing,assaf2026improvisational}. In fact, as Williams has recently argued, social robotics can be viewed as an \textit{Applied Improvisation} project, in which roboticists seek to give robots the same set of skills that theater educators aim to instill in students of improvisation~\cite{williams2025improvising}. However, theatre-based design typically focuses on effectively inducing realistic enactments of interactions that can then be studied, and may have less to say on the specific ways that robot motions should be designed, especially for robots whose morphologies may differ from the human actors engaged in enactments.

\subsubsection{Dance}

In contrast to theater, dance focuses more closely on the design of specific motions. 
Several projects have explored how robots dance~\cite{peng2015robotic,thorn2020human,xia2012autonomous}, while others have explored how robot dance can convey emotion~\cite{bacula2024dancing}. One concrete tool for leveraging dance insights in the design of robot motions is \textit{Laban movement analysis (LMA)}. Proposed by Rudolf Laban, LMA analyzes movements in terms of their direction (direct or indirect), weight (heavy or light), speed (quick or sustained),  and flow (bound or free)~\cite{laban1971mastery}. While LMA was designed to inform the design of human movements in the context of human dance and theater, roboticists have recently shown how it can be effectively used to design emotion-laden robot movements~\cite{lowDOF,humanFormLaban,nonHumanLaban}. 

However, while LMA has seen some use with low-degree-of-freedom robots~\cite{lowDOF,boxDance}, many insights from dance are most clearly applied to dancers with human morphology (as seen, for example, by the high reliance on human morphology in frameworks like the notation used for choreographing with LMA~\cite{robotChoreo}).
Moreover, insights from dance tend to be focused on low-level movement design.

\subsubsection{Animation}
Finally, many insights into robot design have been drawn from 2D animation, which provides concrete insights into the low-level design of specific movements made by non-humanlike characters~\cite{animationLitReview}. In particular, Disney's 12 Animation Principles~\cite{Disney} have informed roboticists on the low level technical details for conveying character behavior~\cite{robotAnimation}. For example, the \textit{Secondary Action} and \textit{Straight Ahead/Pose to Pose} principles have been widely used to shape robot behavior design~\cite{animationLitReview}. \textit{Secondary Action} details how idle animations or other actions that are secondary to a main action can add to the believability of that main action. Terzioglu et al.~\cite{yunus2020cues} explored several animation principles, including Secondary Action, concluding that the appearance of breathing as a secondary action was positively received by human teammates. 
Meanwhile, the principle  \textit{Straight Ahead and Pose to Pose} represent two different ways of controlling an animated character's movement, using either intermediate 'key frames' for a character to hit before continuing the movement, or designing an animated character's movement as moving between pairs of poses. These principles have provided highly concrete philosophies for guiding the practical design of low-level motion plans for robotic characters.
However, most animation principles are less readily applicable in robotics because they intentionally leverage the fact that 2D drawings need not adhere to laws of physics --- a degree of flexibility not afforded to most robots.
The principles of \textit{Squash and Stretch} and \textit{Exaggeration}, for example, are only achievable with highly flexible robots (e.g.,~\cite{sprout24,tofuDraw,ribeiro2012illusion}) unless the principle has been modified for rigid body use (e.g. squashing and stretching movements rather than the body). 

Our examination of these different art-based design approaches suggest several key desiderata and research gaps for robot design. To complement the insights from theater, dance, and animation, there is a need for arts-based insights that (1) can shed light onto the design of physics-constrained characters with potentially non-human morphology, (2) fall at a higher level of abstraction than animation and dance but a lower level than theater, to serve as a bridge between these levels, 
and (3) provide insights into behavior, interaction, \textit{and} morphology design. As we will show in this work, \textit{Puppetry} is an art form that achieves all of these desiderata, and provides a range of insights that naturally complement the insights from theater, dance, and animation.

\subsection{Puppetry}
Like social robot designers, puppeteers manipulate physically embodied characters to convey emotion and stories. Yet it is not yet clear how insights from puppeteering provide \textit{intermediate knowledge}~\cite{lupetti2021designerly,cila2024bridging} for robot designers, in a way that fills the research gaps and desiderata identified above.
Puppetry has been leveraged in HRI to study and enable motion dynamics~\cite{youAsAPuppet, AutonmousTheater, PuppetMaster}, inform robot design~\cite{tofuDraw}, and explore different modes of robot teleoperation~\cite{whatWouldJimHensonDo}. 
However, this prior work has not distilled \textit{intermediate level knowledge}~\cite{lupetti2021designerly} capturing the ways puppetry might generally inform robot  
design.
As described by Cila~\cite{cila2024bridging}, intermediate knowledge like strong concepts~\cite{hook2012strong}, design patterns~\cite{borchers2000pattern}, and design guidelines~\cite{cila2024bridging}, are less general than theories and do not aspire to their breadth~\cite{hook2012strong}, yet are more general than context-specific knowledge, and thus more readily uptaken by scholars, designers, and practitioners.

Therefore, we see puppetry-driven intermediate knowledge as a key research gap. We believe that puppetry-inspired intermediate knowledge has not yet been distilled because of the way puppeteering knowledge has been traditionally recorded and disseminated. For centuries, puppeteering knowledge has passed directly from one puppet master to another --- only recently have puppeteers begun to concertedly document and share practices and resources in print and online. This has resulted in some efforts to present general puppeteering principles, such as Handspring principles, which emphasize the puppeteer's movements and stance. Although important to physical puppeteering, these principles are less applicable to robotics, and do not cover key considerations such as how a puppet talks or displays an emotional state~\cite{handspring}. 
As such, in this work we seek to derive a new set of puppeteering-based intermediate knowledge for robot designers.

\section{Method}
Our research team consisted of one puppeteer, two human robot interaction (HRI) researchers and two undergraduate students. The lead undergraduate student performed an analysis of key texts from the puppetry literature to identify key principles articulated in those texts, and discussed those principles with the other research team members to confirm the accuracy of the principles in the context of modern puppetry practices, and to analyze how those practices might be applied to robotics.

Our analysis of this literature was an ongoing and iterative process throughout a year of research. We began by analyzing two key books~\cite{down,latshaw}, suggested by the team's puppeteer, and three key books~\cite{millar,williams,handspring} suggested by the director of a major puppetry organization. 
These texts were read for insights on how puppets' interaction, behaviors, and interactions are designed. Reoccurring themes were documented and organized with grounding to direct quotes from the text. From the first two texts, an initial set of thirteen puppetry principles were derived. These principles were iterated and expanded on through the insights gained from the additional texts, and from discussion with the team's puppeteer. After this analysis was completed and a tentative set of puppetry principles was derived, the research partner puppeteer reviewed them, gave insights, and iterated on them until the principles successfully represented their expert understanding of current puppetry practices. 


Following the creation of these original thirteen principles, a focus group was conducted with three professional puppeteers, who discussed and evaluated the principles further as they related to their practice. The puppeteers include a professional performer, a puppeteer who focuses on kids' activities, and a puppeteer who runs a local puppeteering company. Together, these puppeteers have over 30 years of experience in puppetry. These Puppeteers' Reviewed each of the thirteen principles and combined or removed principles they felt were unnecessary or repetitive, based on their professional experience in puppeteering. This process resulted in a set of eight core puppetry principles that aim to encapsulate popular modern puppetry practices that can be applied to robot behavior design. 

Then the lead undergraduate drew illustrations of each principle to explain how each principles was meant to be applied to puppets, and developed an explanation of how each principle might be applied to robotics, how it connects to robot design literature, and what novel insights it reveals for robot design.

\section{Design Principles}
In this section, we introduce the eight key puppeteering principles and show their applicability to robot behavior design and literature.  After introducing each puppetry insight, we describe the perspective of puppeteers, gathered through our focus group, on that principle. We then translate that insight into a concrete robot behavior design guideline. And finally, we highlight connections between that robot behavior design guideline and the larger robot design literature. 

\subsection{If the Puppet/Robot is Alive, it is Breathing} \label{s_sec:breath}

\begin{figure}[h!]
    \centering
    \includegraphics[width=.4\linewidth]{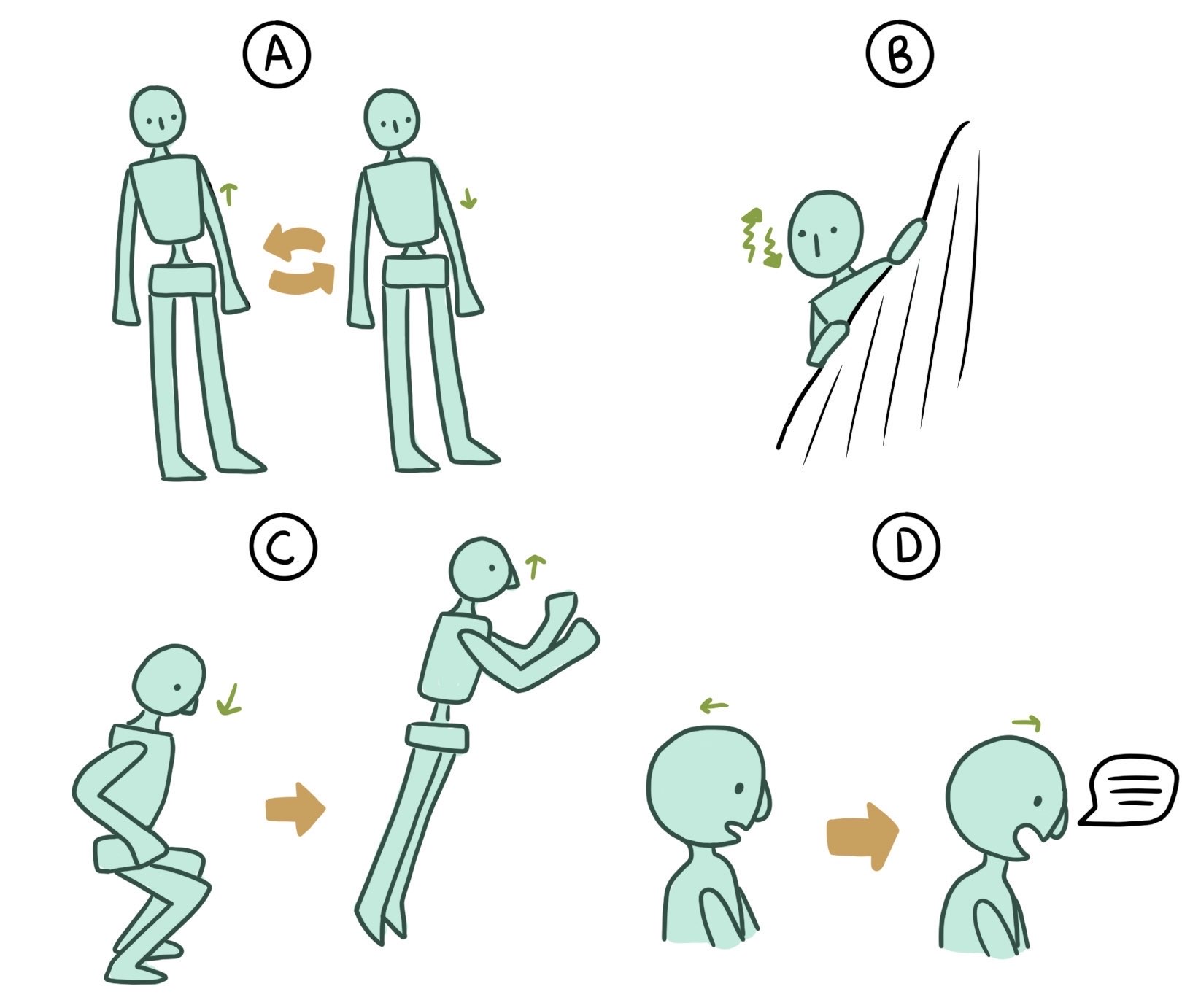}
    \caption{If the puppet is alive, it is breathing: idle breathing motion, breath indicating emotion, breath leading movement, and the Henson-Punch}
    \label{fig:breath}

    \Description{A: Drawings of two frames of a humanoid puppet with its chest moving linearly up and down to signify an idle breath
B: A drawing of a humanoid puppet peeking out behind an object. There are two zig-zagged arrows pointing up and down to indicate its shaky breath. 
C: Drawings of two frames of a humanoid puppet jumping. The first frame is the puppet squatted down before the jump with a downward arrow to indicate an inbreath on the first part of the action. The second frame is the puppet leaping in the air with an upward arrow to indicate the out breath on the release of the action.
D: Drawings of two frames of a humanoid puppet talking. The first frame has the puppet’s head drawn back relative to the body with an arrow to signify an inbreath before speaking. In the second frame, the puppet’s head is drawn forward with an arrow to signify an outward breath and a speech bubble to show that the puppet is talking.}
\end{figure}

One of the main channels used by puppeteers to communicate with their audiences and with each other is \textit{breath}. A puppet is always breathing --- this is true regardless of whether a robot character is anthropomorphic, zoomorphic, or mechanomorphic, but the morphology of the puppet dictates what its breath looks like. Often, breath is portrayed via a slight sinusoidal motion (Fig.~\ref{fig:breath}A), that can be amplified when responding to certain cues. For example, if a puppet is anxious, it might have quick, shaky breaths (Fig.~\ref{fig:breath}B). 
Breath can also be used to initiate movements. When a puppet begins to jump, it might take a quick deep breath inward and an out-breath on the take-off (Fig.~\ref{fig:breath}C).

Our analysis also shows how puppets' breath should mirror their puppeteers' breath, with every in-breath and out-breath reflected by the puppet. This is leveraged by techniques like the Henson-Punch, whereby a puppet's head is drawn back before speaking and then released forward, resulting in the puppet 'spitting' the words out (Fig.~\ref{fig:breath}D). 

\subsubsection{Puppeteers' Review}

In the focus group, all three puppeteers expressed the critical importance of breath as a source of the “life” within the puppet. At the beginning of the focus group, two of the three puppeteers explicitly mentioned, without prompting, that breath was at the core of their practice.

\begin{quote}
    Puppeteer 1: "When breath is the very first thing, like, the first thing we teach people too, so when something's coming to life, whether it's a robot or a puppet the breath is \textbf{always} number one, because you can't do anything else if you're not breathing... That's a hard one for people to conceptualize, of like, you have to breathe in order to start to do something."
\end{quote}

\subsubsection{Design Guideline: If the Robot is Alive, it is Breathing} Robot designers can attend to robots' ``breathing'' behaviors in several contexts: (1) when the robot is portraying a living being, (2) when the robot is expressing emotion, and (3) when the robot is coordinating with others. In each of these contexts, breath can be used in a different way.
When the robot is meant to portray a living being, the appearance of breathing can be achieved through slight sinusoidal moments to its body as a whole or oscillatory elements in the robot's chest. As described above, how the robot "breathes" would depend on its morphology. 
When robots express emotion, the use of breath may change the ways that people interpret those emotions. For example, the excitement and nervousness may manifest differently through changes in breathing patterns. Accordingly, if a robot is meant to be perceived as anxious, a shaky breath pattern may help observers to correctly interpret the robot's emotional state. 
Finally, if a robot is coordinating with others, it may use breath as a medium for non-verbal communication to its partners. A breath inward before an action (e.g., before picking up a box) can signal to human counterparts that the robot is about to act, so that they can coordinate and synchronize their movements. 

\subsubsection{Connections in the Robot Design Literature}
Breath has been previously used in robotics to reduce user stress and anxiety~\cite{Sabison2024breath,asadi2022breath} or to signal emotion~\cite{klausen2022}. However, there is less research on how robots might use breath as an idle behavior~\cite[cf.][]{yunus2020cues}, and none (to our knowledge) on how breath can be used as coordination cues in collaborative activities. 
This represents a key opportunity for future work.

\subsection{The Audience Fills the Gaps}
\label{S_sec:Audience}

In puppetry, the audience plays a key role by interpreting a puppet's behavior in three ways: assuming emotions, directing their attention, and using their imagination to fill in gaps left by a performance. 

\begin{figure}[h!]
    \centering
    \includegraphics[width=.6\linewidth]{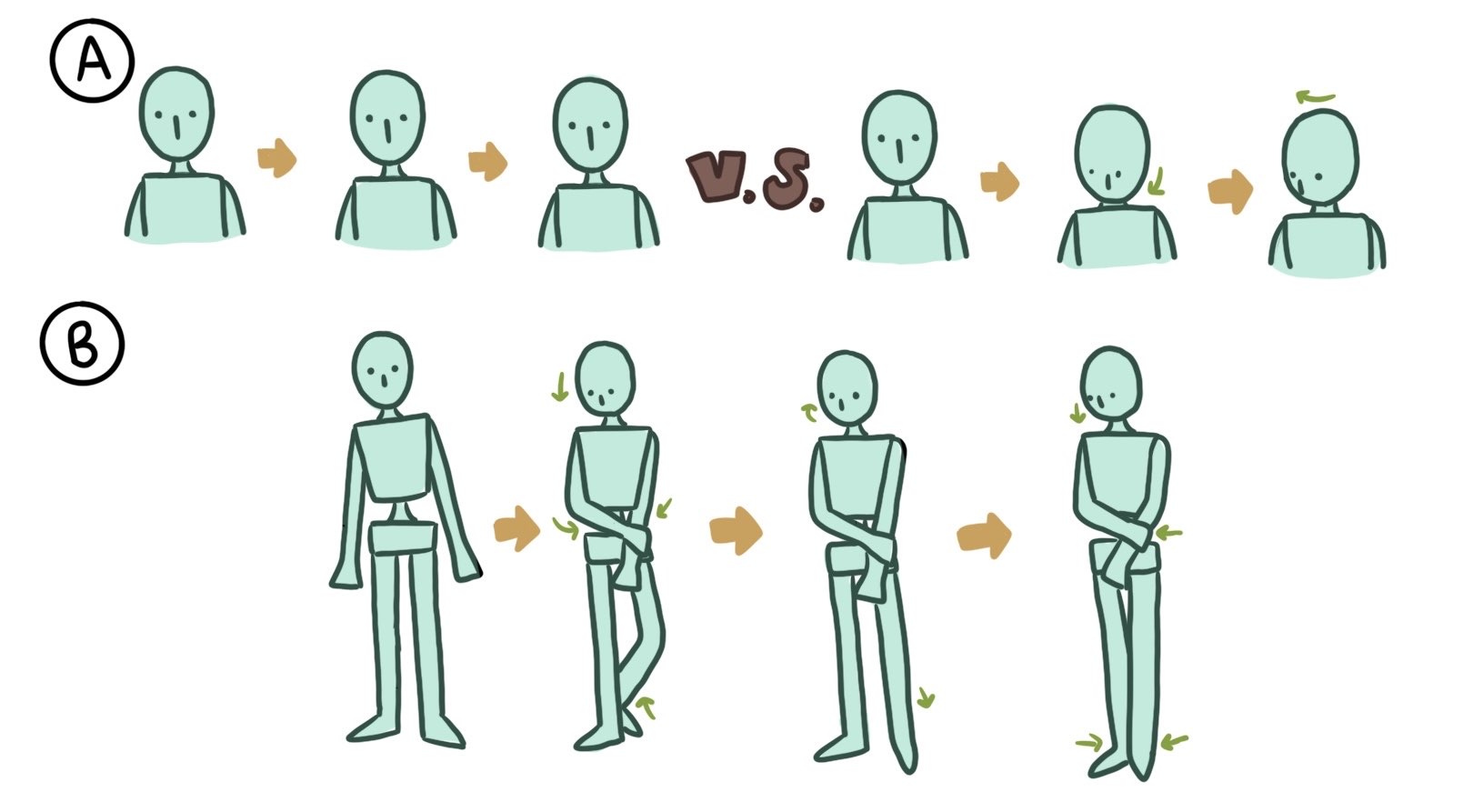}
    \caption{The audience assumes emotion: motion, or lack thereof, communicates emotion and a puppet intentionally showing emotion}
    \label{fig:Puppet Emotion}

    \Description{A:This figure compares two cases side to side. In the first case there are three identical frames where a humanoid puppet is facing and looking forward, intended to give the impression of emotionless staring. The second case contains three frames of the same puppet. The first frame is the same as the one from the last case. In the second frame the puppet moves its head down slightly and in the third frame it turns its head to its right a bit. This movement is intended to give the impression of shyness.
B: This figure has four frames of a full humanoid puppet with arrows indicating direction for every movement that happens between frames. The initial frame is of the puppet standing in a neutral stance with its arms by its side. In the second frame, the puppet looks down, grabs its left arm with its right, and moves its left leg behind its right, closing its body language. In the third frame, the puppet looks up and places its left foot back to a neutral position, thereby slightly opening its body language. The last frame depicts the puppet closing its body language again by pulling its left arm closer to its center, looking back down, and drawing its ankles together.}
\end{figure}

\textbf{Emotion.} The audience will always assume that a puppet has emotion. A puppet's every movement and moment of stillness communicates emotion. Fig.~\ref{fig:Puppet Emotion}A demonstrates how the emotion communicated to the audience can change completely when a small movement is added. Puppeteers take advantage of this phenomenon by constantly manipulating a puppet's emotional state. Puppeteers constantly play into how movements communicate emotion to get the most out of every moment the audience sees a puppet, as shown in Fig.~\ref{fig:Puppet Emotion}B.

\begin{figure}[h!]
    \centering
    \includegraphics[width=.4\linewidth]{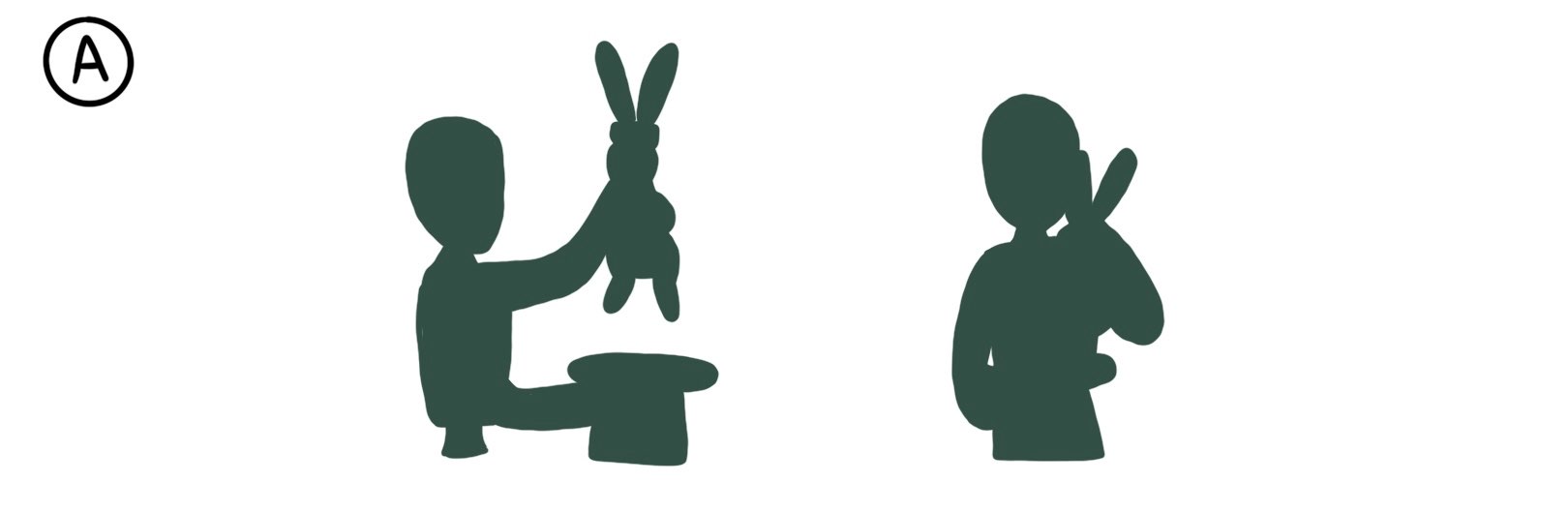}
    \caption{Account for Audience Perception: Movements being led by different body parts, and maintaining a clear silhouette}
    \label{fig:storytelling}

    \Description{A: This figure shows two drawings of silhouettes of a puppet pulling a rabbit out of a top hat. In the first drawing, the puppet is turned to the side and holding its arms away from its chest. The audience can see the silhouette of the top hat and the rabbit, each held in a separate puppet hand. In the second drawing, the puppet is facing the audience and holding the top hat and rabbit in front of it. Since the figure is silhouetted the silhouettes of the individual items are obstructed by the silhouette of the puppet so it is difficult to understand what is happening.}
\end{figure}

\textbf{Attention.} To puppeteer effectively, puppeteers must be aware of --- and account for --- what audiences are able to perceive and where their attention is directed. 
Puppeteers must ensure that the audience can see what the puppet is doing on stage. This requires the puppet to maintain a silhouette while performing actions. Fig.~\ref{fig:storytelling}A shows two silhouetted pictures of someone pulling a rabbit from a hat, the first much clearer than the second. 
In addition, different puppets must not \textit{compete} for attention. While one puppet talks, another must look at it. A gesture towards another puppet when it is its turn to talk indicates to the audience to shift their attention. This technique  is called 'passing the ball' because puppeteers pass the audience's focus around like a ball. 

\begin{figure}[h!]
    \centering
    \includegraphics[width=.4\linewidth]{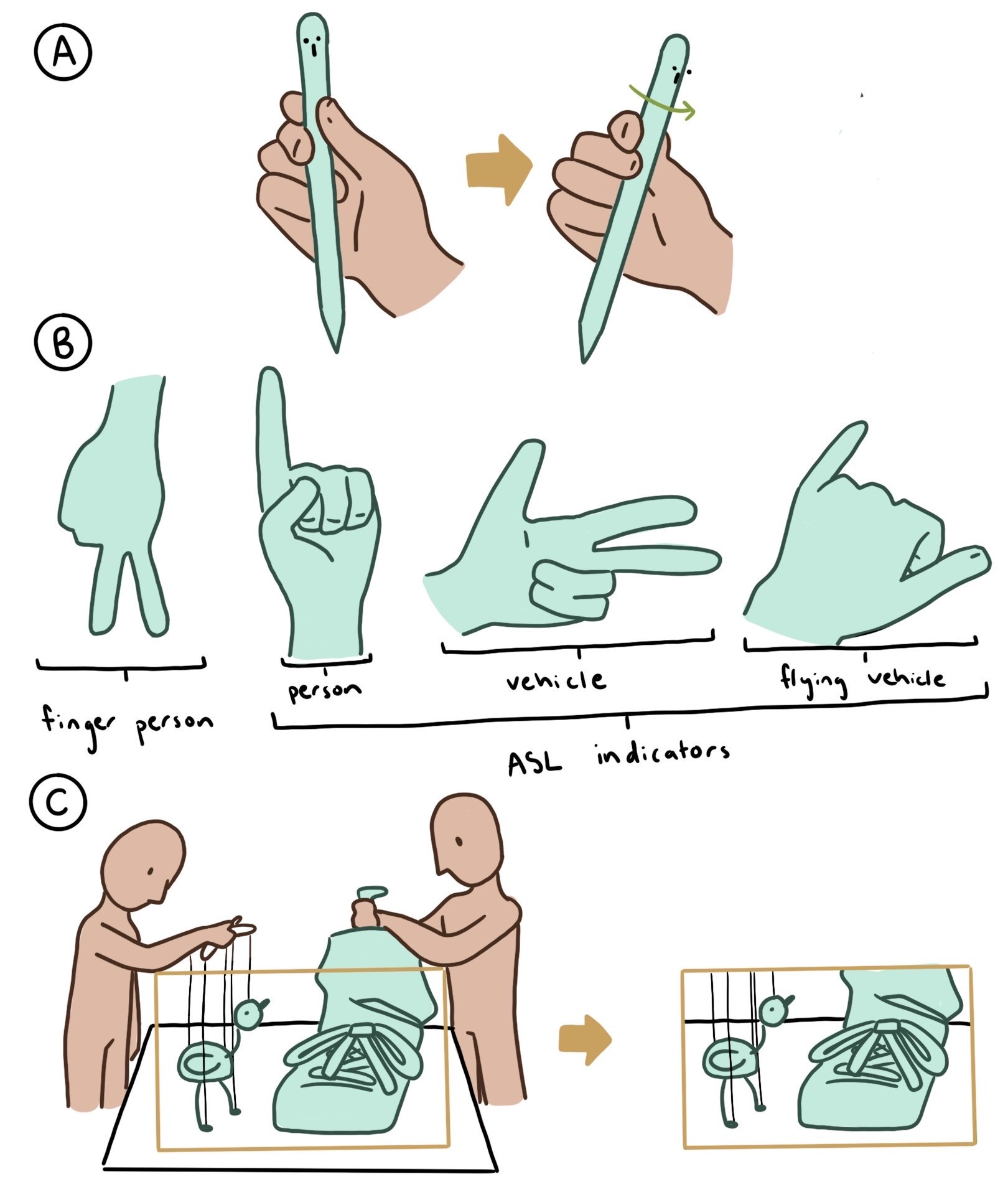}
    \caption{
    Imagination will fill the gaps: hand characters, and controlling the audience's perspective}
    \label{fig:Imagination}

    \Description{A: Two hand drawn frames of someone moving a pencil with a face drawn onto it pinched between the tips of their fingers. In the first frame, the pencil is straight up and then it gets turned to one side and rotated in the second frame, changing the angle and location of the face.
B: Four drawn hands in various positions. The first, labeled ‘finger person’, is a hand pointed downward in a fist  with the pointer and middle finger pointing downward. The next three figures are labeled ‘ASL Indicators’. The first of this category  is labeled ‘person’ and is a fist with the pointer finger pointed upward. The next is labeled ‘vehicle’ and is a hand with the palm facing toward the viewer and the ring and pinky finger folded in so they are touching the palm. The final is labeled as ‘flying vehicle’ and is a fist with the thumb and pinky finger pointed outward.
C: This figure is a drawing of puppeteers puppeteering their puppets for film. The puppeteer on the viewer’s left is holding a bird marionette and the one on the viewer’s right is holding a large foot puppet. There is a yellow box around both the bird puppet and the foot puppet that does not include the top of the foot puppet or the puppeteers. The contents of only this square is to the right of the first drawing to depict what the film’s audience would see when viewing.}
\end{figure}

\textbf{Imagination.} When a puppeteer picks up an object and manipulates it, they can expect that their audience will readily perceive that object as a unique character, envisioning a face and a cone of focus, as shown in Fig.~\ref{fig:Imagination}B. 
This can be seen in the ways that puppeteers use their hands. Fig.~\ref{fig:Imagination}A depicts the many different ways that hands can be used to puppeteer different scenes, such as using one's  fingers to create a ‘finger person’ that can walk around and look at things --- a principle leveraged in highly effective visual communication systems like American Sign Language (ASL). 

\subsubsection{Puppeteers' Review}

While our original literature review suggested three distinct design principles related to imagination, our three puppeteers recommended collapsing these into a single design recommendation with the three \textit{subcomponents} described in this section. Through these three mechanisms, the audience interprets the puppets' behaviors as being intentional, expressing a specific emotion for a specific purpose.

\begin{quote}
    Puppeteer 2 \textbf{(Emotion)}: “It's always intentional, so always treat it with respect and realness.”
\end{quote}

This intention should be clear to the audience, given how movement is interpreted from their perspective. 

\begin{quote}
    Pupeteer 2 \textbf{(Attention)}: "Yeah, I would call this intentional clarity, and accessibility, of just… so that the people on the far back can make sense of what's happening, with movement and other things, but clarity is definitely… Helpful and important to keep in mind.”
\end{quote}

This intentionality also encourages the audience to use their imagination, which can not only bring the puppet to life, but also remove unwanted aspects from the performance, such as set pieces, or even the puppeteer themselves. 

\begin{quote}
    Puppeteer 1 \textbf{(Imagination)}: “Your imagination fills in the gaps, but it also erases things it doesn't need to see, in the sense of when you're on stage with a puppet, the audience will notice you for 30 seconds, and then the puppeteer completely disappears, even though they're right there in front of them. So it's that idea that the imagination can add or subtract.”
\end{quote}

\subsubsection{Design Guideline: Robot Users Will Fill the Gaps\\\nopunct}

\textbf{Emotion.} Robot designers can use affective displays to head off users' misconceptions about robots' emotional states.
Humans naturally anthropomorphize objects~\cite{heider1944experimental}, and this effect is greater when objects are self-propelled or animated~\cite{barrett2003inanimate,fink2012}.
For example, if a robot is calibrating, an observer may not understand the reasons for its movements. 
Although its actions may be more aligned with stretching or fidgeting, observers might interpret its motion as a shrug of apathy. 

 To prevent possible misconceptions about robots' emotional states, roboticists should intentionally attend to the emotional states that they want audiences to interpret, and use consistent affective displays aligned with those states. This may be especially important not only during standard usage, but also during less common behavior modes, such as calibration and shutdown.

 \textbf{Attention.} Just as puppeteers must ensure that puppets are angled in such a way as to be optimally interpretable, so too must roboticists design their robots' physical behaviors to ensure that they can be interpreted.  As in puppetry, this can be achieved by ``cheating out'' or, when the robot can only be seen from certain angles, through the use of anticipatory cues such as ``passing the ball''.
Similarly, just as puppeteers must ensure that when one puppet is talking, other puppets are not distracting, so too might roboticists, when one robot is talking, strategically use stillness to reduce the risk of attention being lost from the speaking robot.

 \textbf{Imagination.} While robot designers often use abstract robot morphologies with minimal surface details~\cite{phillips2018human}, robot designers might go further through the use of framing to suggest morphological features that cannot be reasonably constructed. As an example, we consider novelty piggy banks with ostensible animals inside that "reach out" to grab coins --- a design that requires the toy constructors to only build the parts of the character that will be seen by the user. Similarly, a robot designer might design a robot that, when it reaches behind the desk, a hand appears elsewhere in the room and grab something, prompting the user to imagine an invisible arm stretching out (behind the walls) to link robot hand and torso.

 \subsubsection{Connections in the Robot Design Literature\\\nopunct}
 
\textbf{Emotion.} There has been substantial research in the HRI literature into the design of effective affective displays through facial expressions~\cite{Danev2017expression,park2015expression} or body language~\cite{McColl2014bodylanguage,embgen2012bodylanguage,knight2014}. Anthropomorphism of robots is also well researched~\cite{fink2012, spatola2021anthro}. There is also some research into how idling movements may effect a robot's perceived emotion~\cite{Cuijpers2015} and how users may misinterpret emotional displays~\cite{Pelikan2020}.
However, there is less research into the emotional states conveyed through idle animations, the intentional use of affective displays to avoid emotional misconstrual, or how robot overreaction may impact how users interpret movement. These areas thus represent key directions for future work.

\textbf{Attention.} There has been some very limited work on robots angling their bodies for optimal communication~\cite{mead2017autonomous,el2017real}, including some recent work on predicting visual differentiability~\cite{fletcher2022predicting}, none of which appears to take a performance-driven perspective. This suggests a key opportunity for future work. 

In contrast, there has been substantial literature on interaction shaping~\cite{gillet2024interaction,matsuyama2015four}, especially re-directing attention using gaze cues~\cite{Yamazaki2008turntaking,gillet2021robot,scassellati2018improving}, although this work primarily assumes a single robot interactant. There has been some work on communication from multi-robot systems~\cite{bejarano2022you,bejarano2023no,Williams2021}, and some work on inter-robot gaze~\cite{hayashi2008robot}, some of which takes a puppetry-oriented perspective~\cite{sanoubari2023using}. However, none of that work has strategically leveraged stillness, thus presenting another key opportunity for future work.

\textbf{Imagination.} While selective embodiment is prominent in animatronics (e.g., at theme parks), there has been little use of this strategy in the HRI literature. The closest work we are aware of is work on Mixed Reality Deictic Gestures in which armless robots are given virtual hands that can point to referents~\cite{zhu2024designing}. However, these gestures are not designed in a way that expects users to infer elements of morphology that are not truly there. This suggests that \textit{selective embodiment} may be a key unexplored tool for robot designers.

\subsection{Use Movement and Stillness Carefully}
\begin{figure}[h!]
    \centering
    \includegraphics[width=.4\linewidth]{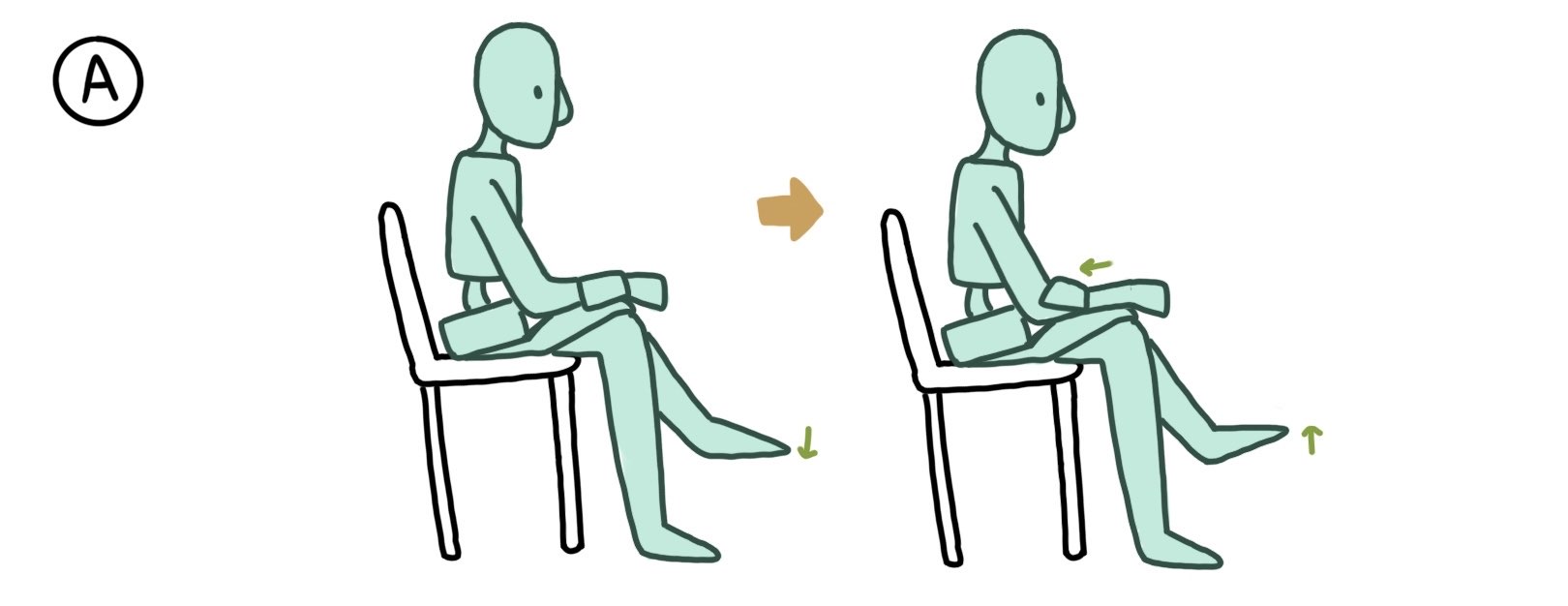}
    \caption{Use movement and stillness carefully: subconscious movement}
    \label{fig:moving puppets}

    \Description{A: There are two frames of a humanoid puppet sitting in a chair with its legs crossed with the left arm resting on its leg. In the first frame, its right hand rests on its left wrist and its foot that is in the air is moving down. In the second frame the puppet moves its right hand up its fore arm and moves its foot upward to insinuate that the puppet is bouncing its leg.}
\end{figure}

Even beyond breath, living things are constantly moving. People make subconscious movements to shift their weight, fidget, or gesture.
The feet of a puppet demonstrates how robots must move in just the same way (Fig.~\ref{fig:moving puppets}A). The feet move even when this is not given any thought, which often leads to them being a puppet's `tell' for its emotional state. They drag when sad, bounce when happy, and fidget when anxious. When the feet are still, they are still for a reason, whether to provide stability or to convey focus.
As a result, each movement of the puppet is a powerful indicator of the robot's emotion, regardless of whether the character depicted by the robot is anthropomorphic (see also Sec.~\ref{S_sec:Audience}). As such, while a puppet should (almost) always be moving,  the typical movements of a puppet should be small, only moving one or two body parts at a time. Moreover, the need for near-constant movement makes moments of stillness powerful. When a puppet is still, it captures the attention of the audience and holds them in suspense.

\subsubsection{Puppeteers' Review}

All of the professional puppeteers agreed that it is not just movement that is critical, but also the balance with stillness in puppetry. They further related this to the subconscious movements that living beings perform and to how these movements maintain a lifelike quality, even when the puppet is not the main focus of the scene. 

\begin{quote}
    Puppeteer 2: "That is part of life, of existing within yourself, and existing within your environment, of having that subconscious movement, being like, oh, what's that over there? I got distracted. Oh, right, I'm returning. Focus"
\end{quote}

\begin{quote}
    Puppeteer 1: "They're doing these subconscious things, that the puppeteer might not even be controlling necessarily, but it's these little subtle things that are keeping them alive, especially when they're not the focus of a scene, or the focus of the action, or whatever."
\end{quote}

\subsubsection{Design Guideline: Robots Must Use Movement and Stillness Carefully}
Robot designers can use movement and stillness to convey emotion and significance to robot behavior. Robots should move in ways that are organic to their morphology, constantly moving one or two features, while using stillness strategically. 
When robots default to idle animations that involve small constant movement, stillness may be especially powerful. In such cases, stillness may communicate focus, or a change in thought process. 
With the power of stillness in mind, robot designers might carefully consider how they sequence movement and stillness. For example, while a robot might use idle animations outside of conversation to convey availability for interaction, it might de-emphasize those animations during conversation (excepting communicative gestures) to convey its focus on the conversation.

\subsubsection{Connections in the Robot Design Literature}
While idle animations have previously been explored in robotics~\cite{Asselborn2017,Song2009,Mutlu2011}, we are not aware of prior work intentionally exploring stillness as a means of conveying focus. This thus represents a key opportunity for future work.

\subsection{The Puppet/Robot Must Focus on Something}
\begin{figure}[h!]
    \centering
    \includegraphics[width=.4\linewidth]{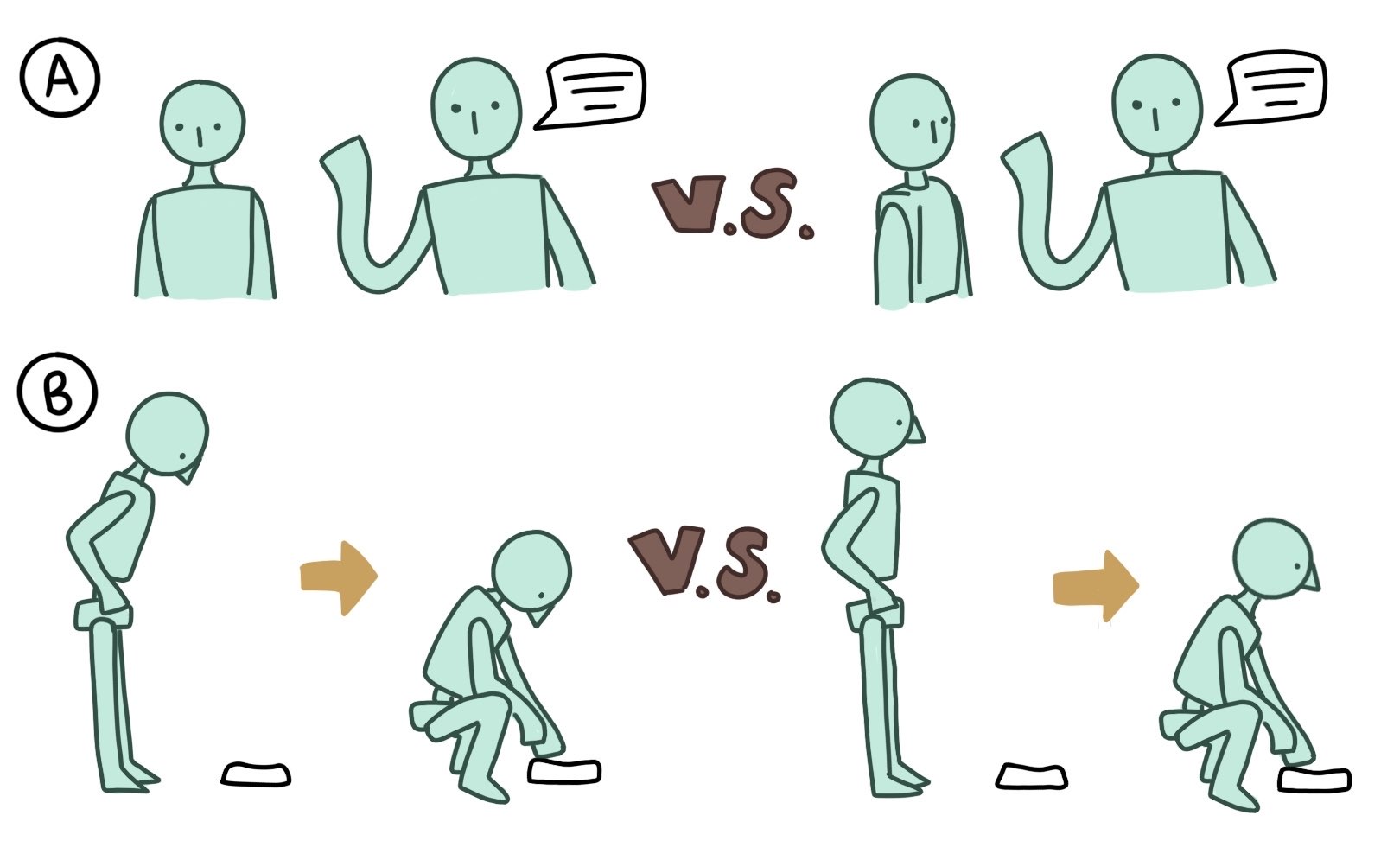}
    \caption{The puppet must always be focused on something: Focus indicating interest and communicating intent}
    \label{fig:puppet focus}

    \Description{A: This figure has two cases compared side by side both with two humanoid puppets standing next to each other. The puppet on the viewer's right is raising its hand and has a speech bubble next to it to signify that it is the talking. In the first case the second puppet facing straight towards the viewer, and turned slightly toward the speaking puppet in the second  case.
B: This figure has two cases of a humanoid picking up an object from the ground compared side to side. In the first case, the puppet looks at the object before picking it up (first frame), and continues to look at the object when it squats down to pick it up (second frame). In the second case, the puppet stares straight ahead before picking up the object (first frame), and continues to stare forward when it squats down to pick it up (second frame).
}
\end{figure}

 The focus of the puppet indicates its interest. As a result, as shown in Fig.~\ref{fig:puppet focus}A, a puppet's focus can indicate who is a part of a conversation. This is true for any robot whose design indicates a deictic cone of focus. If two puppets are talking to the audience, their focus should be directed towards them. If instead the puppets are talking to each other and the audience is only listening to them, the puppet's focus should be directed to their counterpart, rather than the audience. 
The combination of a puppet's focus and intention can communicate what the puppet will be doing next. When a puppet goes to pick something up, it is clear to the audience what its intention is if it looks at the object, takes a breath, and then picks it up. If the puppet does not look at the object before picking it up, it may appear distracted from its task, as shown in Fig.~\ref{fig:puppet focus}B. 
As a result, puppeteers use techniques like  the 'Magic Triangle', whereby a character slightly crosses its eyes, to help users understand where it is looking. 

\subsubsection{Puppeteers' Review}

The professional puppeteers indicated that controlling the puppet's focus is critical for signaling the next movement to the audience and important for allowing them to interpret the behavior as being intentional. They pointed to this focus in their puppetry through a process of “see, feel, react.”

\begin{quote}
    Puppeteer 2: “Yeah, and a similar thing, I usually, play with \textbf{see, feel, react}. So, having that emotional connection of, like, you notice something, and then you have an emotional response to it, and then you react.”
\end{quote}

\subsubsection{Design Guideline: The Robot Must Focus on Something}
Just as a puppet's gaze needs to remain directed, so too should a robot's. This is critical even if a robot does not have visible eyes~\cite{szafir2015communicating}. 
%
Gaze is a tool for communicating intent, interest, and directing focus, however it is important to consider how the robot is using gaze and in what context. If a robot's focus is not directed, users may still assume that the robot is looking at something or scanning the area. 
For example, if a robot is blankly looking forward during an interaction, users may assume that the robot is not invested in the conversation. To have better control over what users assume about the robot, robot designers must choose what the robot would be looking at (e.g. eye contact during interactions).

\subsubsection{Connections in the Robot Design Literature}
Gaze has been a highly explored topic in the Human-Robot Interaction literature.
Gaze is a key tool and anticipatory cue that robots can use to direct user attention~\cite{knight2013gaze} in order to establish joint attention~\cite{Kiilavuori2021eyecontact,Xu2016gaze,Kompatsiari2019} and thus improve task performance~\cite{Huang2010} and human-robot collaboration~\cite{Psarakis2025cues, Mutlu2012gaze}. Without these uses of directed gaze, these types of interactions will falter. Human-like gaze model~\cite{Mutlu2012gaze,Briggs2022gaze} represent a key way tool for ensuring that robot gaze is under constant modulation, and is responsive to both bottom-up and top-down cues. More research is needed into users' perceptions of robots that use such models.

\subsection{When the Puppet/Robot Talks, it Moves Accordingly}
\begin{figure}[h!]
    \centering
    \includegraphics[width=.4\linewidth]{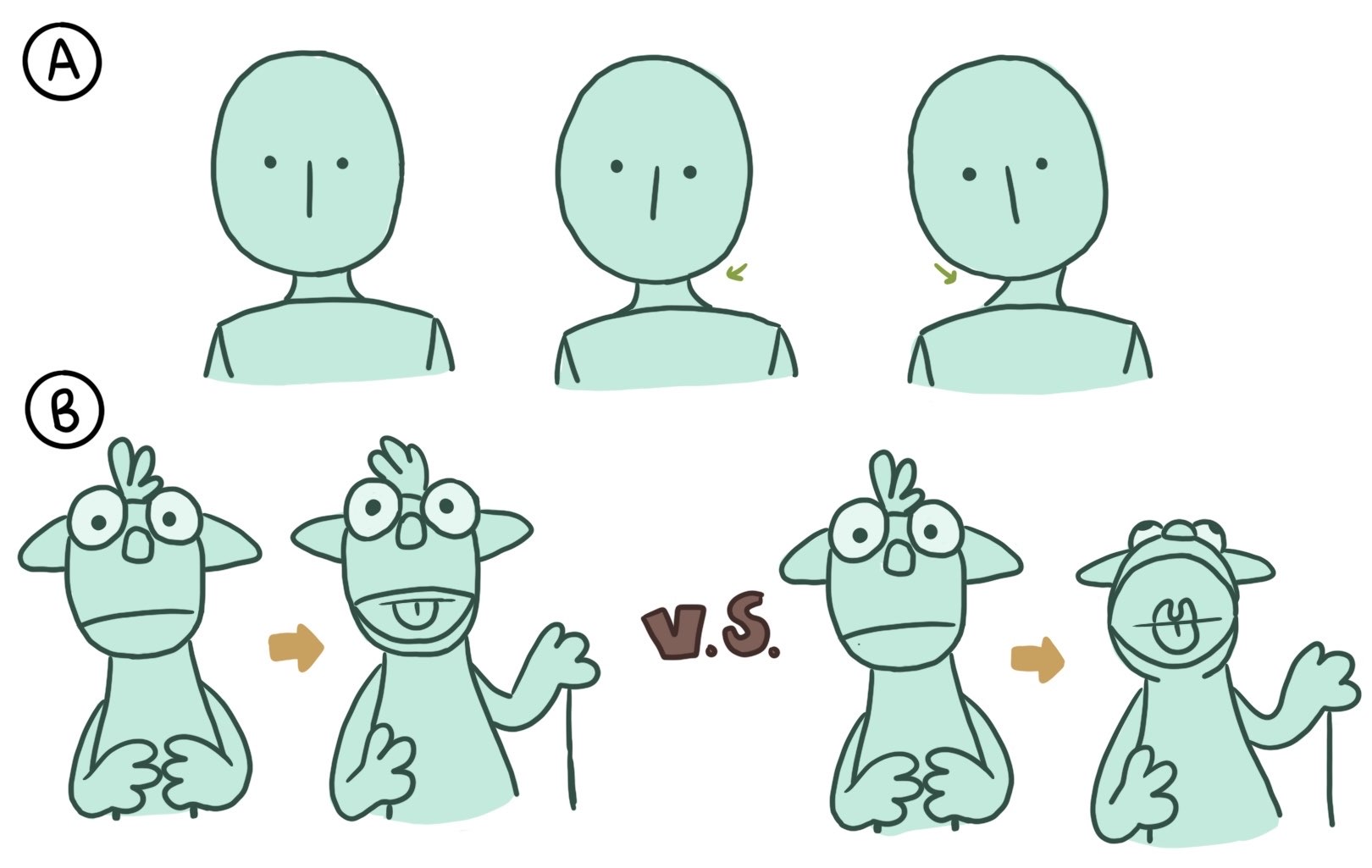}
    \caption{When a puppet talks it moves accordingly: How a puppet without a mouth might indicate speech and keeping focus while speaking}
    \label{fig:talk}

    \Description{A: This figure has three frames of a headshot of a humanoid puppet. The first frame has the puppet head looking straight forward, and the second and third frame has the head tilted left and right respectively.
B: There are two cases of a creature-like hand and rod style puppet talking, both starting with the same frame of the puppet looking forward, mouth closed, and with its hands folded in towards its chest's center. The second frame in the first case depicts the puppet opening its jaw by bringing its bottom jaw downward, keeping its eyes in relatively the same position so it is still making eye contact with the audience. The second frame in the second case has the puppet open its mouth by looking upward, breaking eye contact with the audience. Both second frames have the puppets left arm move out and up in a general gesture.}
\end{figure}

The morphology of a puppet (including its degrees of freedom, means of control, and facial structure) all determine how it talks.
When a glove (or hand) puppet talks, it gestures to indicate that it is the one talking. Speech does not make the puppet bounce up and down because speech does not cause a human's entire body to bounce. In contrast, Bunraku-style puppets, which have flexible necks and no movable mouths, speak by 'waggling' the bottom jaw as if it is getting hit by spoken consonants (Fig.~\ref{fig:talk}A). Finally, hand-and-rod puppets, which often have movable mouths, add the element of lip sync. While puppeteers use different techniques for lip sync, it is universally important to ensure the mouth closes on each constant. It is also important to avoid (during normal conversation) having a 'flip-top lid' in which the top of the puppet's head moves as it talks, as this breaks the puppet's focus (Fig.~\ref{fig:talk}B). 

\subsubsection{Puppeteers' Review}
The puppeteers emphasized the importance of moving while speaking, in accordance with the emotion the puppet is expressing, who it is speaking to, and the dialogue. 

\begin{quote}
    Puppeteer 2: “But I do, I like the way that you word it, with the, it moves accordingly.
    Because, like, it does, like, if it's screaming, it would look more like that last one [Fig. 7b-2], and if it's just having a chat, it's gonna look like that first one. So I think accordingly covers that really succinctly.”
\end{quote}

It was also discussed that this movement may change based on the puppet's morphology, as some may not have a typical open-flap mouth and instead rely on other aspects of their design to indicate the subtle movements associated with speaking. 

\begin{quote}
    Puppeteer 1: "Whatever the text is is being matched as closely as possible in the puppet, whether that is a manipulation of a mouth, or a manipulation of the head, or the whole body, but it's following the cadence of the speech of the text."
\end{quote}

\subsubsection{Design Guideline: When the Robot Talks, It Moves Accordingly}

Robot designers should be sure to accompany robots' speech with co-speech indicators, based on the robot's morphology. Robots with movable mouths may move them in ways that are similar to puppets, while others may use glowing lights; robots with arms may use them to gesture, while armless robots may use body movements like bouncing. 
Co-speech indicators may be important not only to make a robot's speech seem more natural, but also to help distinguish between robots within groups. 

\subsubsection{Connections in the Robot Design Literature}
Robot designers often use co-speech gestures to enhance user understanding or recall and facilitate perceptions of robot animacy~\cite{Meena2012gesture,Huang2013gesture}. This includes research on robot head movement to match utterances~\cite{Ishi2010nodding}.  It has also been suggested that head nodding and gestures can help facilitate turn taking~\cite{Meena2012gesture}, which is described more in Sec~\ref{S_sec:Audience}. Some robots are capable of lip sync through articulated or digital mouths~\cite{Hyung2016, Abdelaziz2025}. Having robot with humanoid features lip-sync may increase the naturalness and clarity of speech.
Non-humanoid robots have indicated speech through lights~\cite{kobayashi2011blinkinglights, Whittaker2021} and whole body movement~\cite{Williams2021}. The latter helped participants decipher which robot in a robot group is speaking.

\subsection{The Puppet/Robot Responds Truthfully to its Given Circumstances}
\begin{figure}[h!]
    \centering
    \includegraphics[width=.4\linewidth]{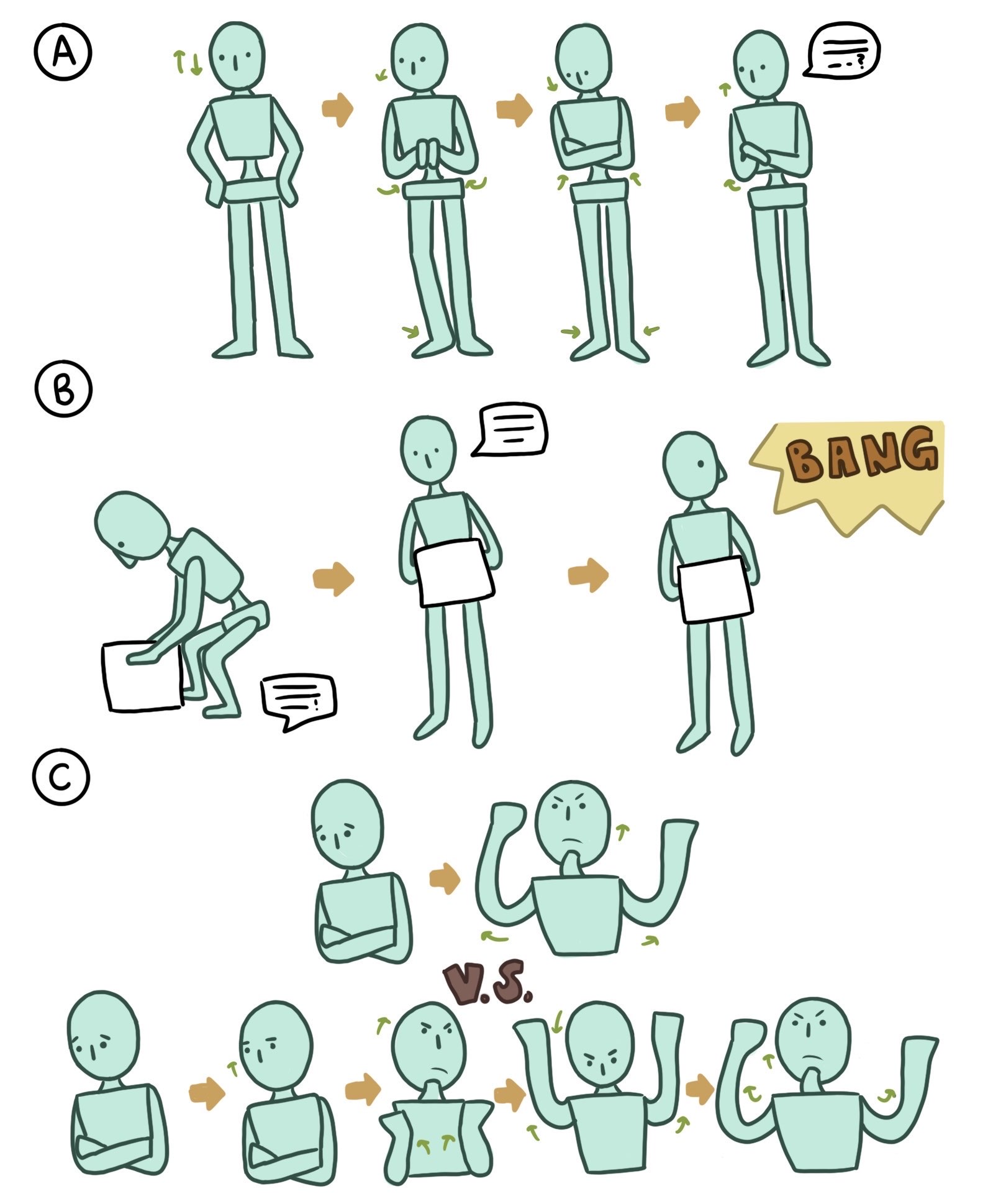}
    \caption{The puppet responds truthfully to its given circumstances: thoughts visibly progressing before taking an action, a puppet responding to cues in its environment, and emotion transitioning from one state to another}
    \label{fig:given circumstances}

    \Description{A: There are four frames of a humanoid puppet with its whole body showing. The first frame has the puppet standing in a confident pose, hands on hips, and nodding, presumably to whoever is talking to it. In the second frame it moves its head down slightly, still making eye contact with the audience, clasps its hands together in front of its chest, and shifts its feet. In the third frame, the puppet looks down, breaking eye contact, crosses its arm, and moves its ankles in towards each other. In the fourth and final frame, it looks back up, making eye contact with the audience, gestures one hand forward, and asks a question (signified by a speech bubble with a question mark).
B: This figure has three frames depicting a puppet picking up a box. In the first frame the puppet is being asked a question by a person off screen while it is squatting down and about to pick up the box. In the second frame, the puppet has turned toward the person while holding the box, and responds. In the third frame, the puppet hears a loud bang noise and turns to look at what caused it.
C:There are two cases in this figure compared vertically with a humanoid puppet depicted from the chest up, including arms. The first case has two frames, while the second has five. In the first case, the first frame starts with the puppet looking to one side with its eyebrows arched and its arms crossed, looking concerned. In the second frame, the puppet has thrown its arms up into the air and is looking at the sky angrily. The second case starts with the same frame as the first case, with the puppet looking off to one side concerned. The puppet looks up, eyebrows furrowed in the second frame. In the third frame the puppet turns its head to the other side and upward, draws its eyebrows in so it looks angry, and uncrosses its arms and flips its wrists upward. In the fourth frame the puppet looks downward, still angry, and throws its arm up so the tips of its hand are aligned with the top of its head. The fifth and final frame is the same as the last frame of the first case, where the puppet draws its hands slightly down and out, and tilts its head up towards the sky.}
\end{figure}

The way a puppet responds to different environmental cues will depend on its internal state, or its thoughts. An example in Fig.~\ref{fig:given circumstances}A shows that if a puppet is responding to someone speaking about a confusing subject, the puppet will show signs of being confused before asking a question. The puppet in the figure appears hesitant or shy to ask questions, revealing its unconfident internal state.
Fig.~\ref{fig:given circumstances}B shows another example, in which a puppet responds to a question, picks up a box, and reacts to a loud noise in the distance. How this puppet could have reacted depends on its internal state. If the puppet was angry and its role was managing those around it, it could have snapped at the person asking a question and then yelled at whoever made the noise. Instead, the puppet in the example responded calmly, reflecting its focused mindset.

It is important to note that when puppets are responding to their circumstances, their emotions may build on themselves or change abruptly. Fig.~\ref{fig:given circumstances}C shows the difference. The first version may communicate a silent snapping point for the puppet, while the second may communicate a dramatic building of emotion. Each start and end at the same key frames, but reflect different internal states.
\subsubsection{Puppeteer Review}

As a principle, responding adequately and as expected given the puppet's circumstances was discussed as core to the puppet's ability to be perceived as alive and aware. The puppeteers related it back to how living beings exist in the world and respond to aspects of their environment in real time.

\begin{quote}
    Puppeteer 1: “I would say, like, as a core principle, yeah, the puppet responds truthfully to its given circumstances. I call this world being… the puppet is just being in the world, it’s just existing… just how we exist... So yeah, I think that this is definitely a core emotional and physical [principle].”
\end{quote}

\begin{quote}
    Puppeteer 2: “Yes, ... that is the character's truth, that is performing it honestly, that if there is a bang in the audience during a show, it's not necessarily a scripted moment, but if it is in the moment, it recognizes that, or adjusts accordingly.”
\end{quote}

\subsubsection{Design Guideline: The Robot Responds Truthfully to its Given Circumstances}
Robot designers should design their robots to react to external stimuli. If a robot does not respond to a cue (e.g., a siren, being touched, etc.) in the way expected by its audience, this may lead to distrust and decreased perceptions of competence, intelligence, and animacy. To make a robot appear grounded in reality in the same way humans are, designers should consider the external stimuli a robot is likely to face, ensure the robot is able to detect and characterize those stimuli (with sensitivity to privacy and other ethical considerations), and either (1) ensure the robot is able to respond to them, (2) design the robot so that it is excusable if it does not respond to them, or (3) design the robot in a way that prevents or discourages stimuli to which the robot would not be able to appropriately respond.

\subsubsection{Connections in the Robot Design Literature}
Robots naturally require sensors to perceive and interpret their environments; and moreover, robot actions or reactions are contextualized by outside stimuli and previous actions~\cite{Pelikan2020, Francis2025socialrobot}.
Robot perception is also critical from a theatrical perspective: Williams argues that social robotics is inherently an Applied Improvisation project~\cite{Williams2025improv}, and highlights the importance of \textit{hyperperception} to effective human and robot improvisation~\cite{pelletier2017production}.

Some social robot designers choose to incorporate vision sensors for user, emotion, and gaze detection to trigger different responses~\cite{Hoffman2015reaction,Hoffman2010jazz,Palinko2016gaze}. Robot designers also use pressure sensors allow robots to gauge how to handle certain objects and respond to touch cues~\cite{Savanovic2015paro, Bethel2018therabot, pan2022pressure}.

There has been some research we might consider research on low-latency hyperperception, including work on social robot navigation~\cite{Francis2025socialrobot}, robot participation in improvisational jazz~\cite{Hoffman2010jazz}, and robot response to expected touch~\cite{Bethel2018therabot,Savanovic2015paro}.
And, there has been some research on robot detection of anomalous or surprising stimuli~\cite{white2014surprise,park2016multimodal,park2019multimodal}, albeit not usually with respect to unexpected stimuli in the way described above. Moreover, there has been little research onto robots' affective responses to surprising stimuli, with the possible exception of recent work on robot response to adverse behaviors~\cite{du2024can,Hoffman2015reaction}. Accordingly, this represents a key open area for future work.

\subsection{Control the Puppet's/Robot's Weight; Weight can Convey Emotion}
\begin{figure}[h!]
    \centering
    \includegraphics[width=.4\linewidth]{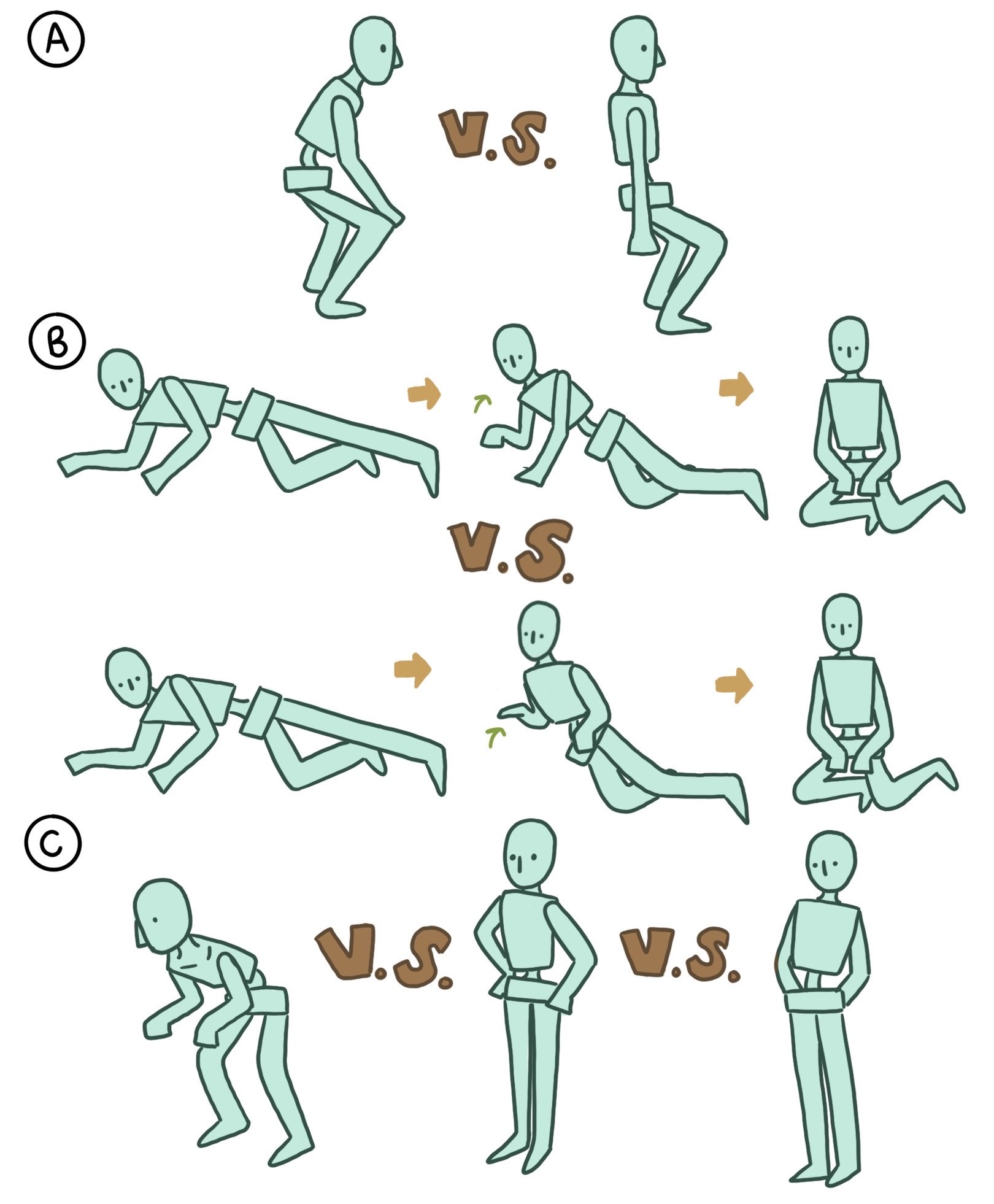}
    \caption{Control the Puppet's Weight: posture conveying balance, movements grounded in gravity, and posture communicating personality}
    \label{fig:weight}

    \Description{A: There are two cases compared side by side. In the first case a humanoid puppet is crouched, leaning forward so the top of its chest is aligned vertically with the center of its feet. It also rests its hands on its knees. In the second case, the puppet is squatted down, with its torso completely vertical so its chest is vertically behind the heels of its feet. In this case, the arms are dangling straight down.
B: In this figure there are two cases compared vertically of a humanoid puppet getting up from lying down. In both cases the first frame is of the puppet lying on the ground, and the last (third) is of it sitting on the floor. The intermediate frame depicts the puppet sitting up. In the first case the puppet places its arm on the ground as if to support itself. In the second case the puppet does not place any hands down, giving the effect that it is unbalanced.
C: This figure compares three different postures on the same hand drawn humanoid puppet. The first posture has the puppet crouched down so its torso is nearly horizontal, with its arms bent with its hand drooping down so they are above its knee. The second posture has the puppet with its chest puffed outward, feet drawn close together, and head firmly looking forward. The third posture has the puppet’s chest drawn back relative to its pelvis, effectively sticking out its stomach, and its hands clasped behind its back.}
\end{figure}

Puppeteers maintain the center of gravity of a puppet to maintain balance and communicate personality. While puppets are subject to the laws of physics, they are typically supported by external forces like rods, hands, and strings, which can allow them to appear unaffected by gravity. Fig.~\ref{fig:weight}A shows how something as simple as the slope between a puppet's back and hands can change the perceived balance of the character. Fig.~\ref{fig:weight}B shows the difference between a puppet getting up from laying down when it uses its hands as an anchor and if the puppeteers skipped that step and just allowed the puppet to rise without any gravitational effects.
Finally, puppeteers use posture to convey  personality and emotional state. The different ways a character carries itself, or what body part it leads with when moving, can convey different personalities. Fig.~\ref{fig:weight}C shows three different characters, each with a different posture, inspired by the distinctive characters in Commedia dell'Arte.

\subsubsection{Puppeteers' Review}

Beyond posture, the professional puppeteers mentioned considering the entirety of the puppet's form, including any limbs it may have in its morphology. They include the importance of weight in conveying emotion across all aspects of the puppet's morphology and movement, calling this aspect “emotional indicators.” 

\begin{quote}
    Puppeteer 1: “A term that you might use is emotional indicators. So, something that's on a puppet's body, including posture, including how they're carrying themselves, are emotional indicators. So, like, a really good example is, like, a dog, like, a dog's tail is an indicator, because it tells us how it's feeling, like, it's a physical part of their body... But it's like, there's physical things on the puppet that are going to express emotion, one of them being posture, like you just said.”
\end{quote}

\subsubsection{Design Guideline: Control the Robot's Weight; Weight Can Convey Emotion}

While puppets are typically supported by external forces such as hands and strings, robots are typically self-supported characters. Nevertheless, roboticists might still manipulate how a robot manipulates its center of gravity or holds itself, either as a channel for implicit communication~\cite{ju2008design} (e.g., to signify that the robot is under strain) or to communicate the robot's personality. 

\subsubsection{Connections in the Robot Design Literature}
There has been some prior work on robot posture as a means of conveying status~\cite{stein2022power} and emotion~\cite{briggs2014robots}, as well as research on robots whose expressivity comes from contortion of their body shape~\cite{sprout24}. There has also been some research into manipulation of the posture of robotic rats~\cite{shi2018modified}, albeit in the context of rat-robot rather than human-robot interaction studies. However, these works typically focus on changes in posture moment-by-moment to convey state; more research is needed into the use of posture to consistently convey personality across longitudinal human-robot interactions.

\subsection{There is an Economy of Movement; Avoid Overemphasizing Details}

\begin{figure}[h!]
    \centering
    \includegraphics[width=.4\linewidth]{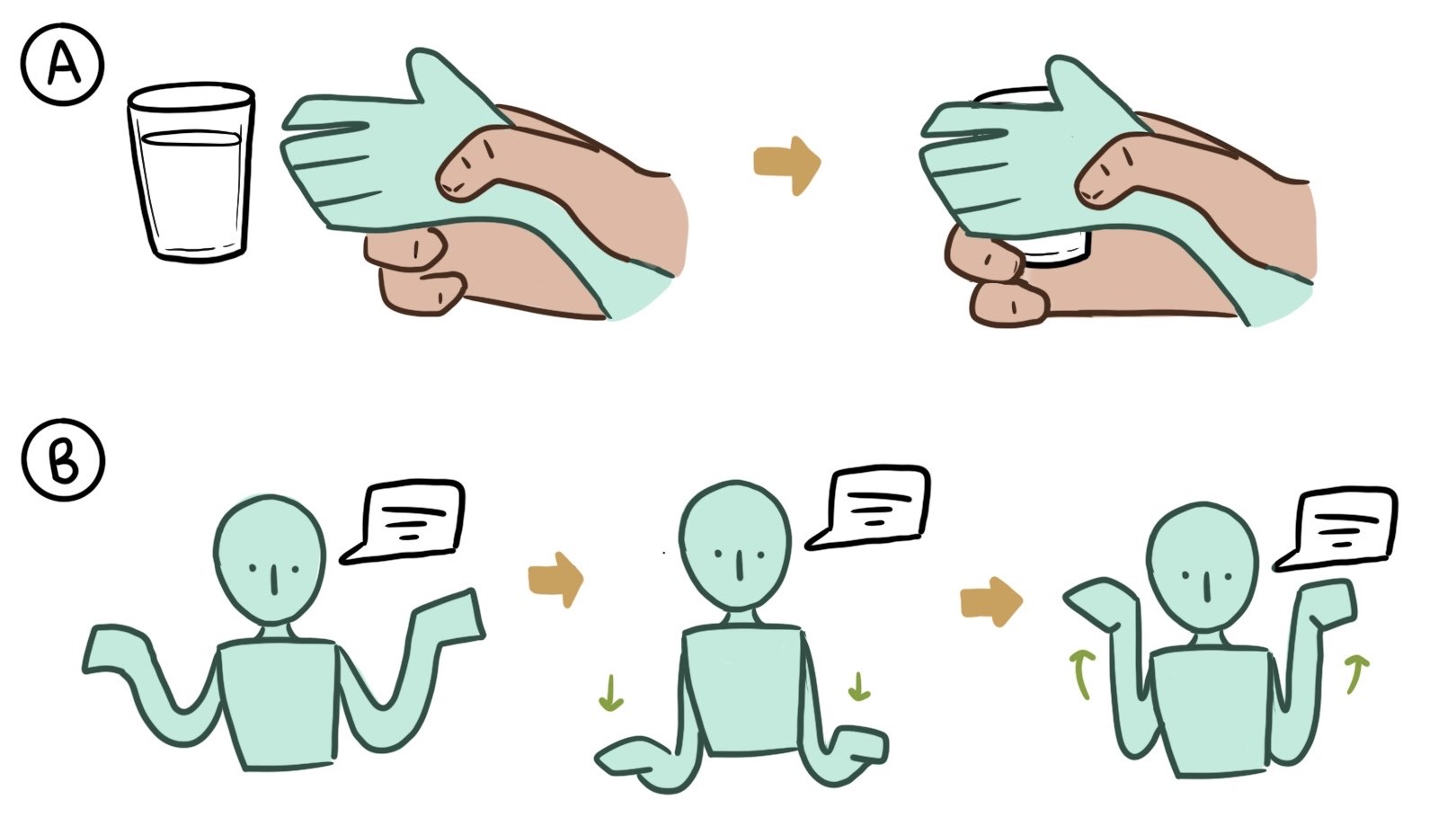}
    \caption{Avoid Overemphasizing Details: the puppet does not need to be capable of intricate functions if the function can be implied, and over-emoting a casual conversation}
    \label{fig:details}

    \Description{A: This figure has two frames depicting a puppeteer making a puppet ‘pick up’ a prop cup. The first frame has the cup next to a puppeteer's hand pinching a puppet’s wrist between their thumb and palm so that their fingers are behind the puppets hand. In the second frame the puppeteer is holding the cup behind the puppets hand, and the puppet’s hand has not changed shape.
B: This figure has three frames depicting a humanoid puppet’s top half while it is speaking and gesturing. In the first frame, the puppets arms are open and its hands are faced upward and slightly above shoulder level. In the second frame, the puppet draws its arms straight down so its hands are at the bottom of its chest. The puppet then draws its straight hands upward to the middle of the puppet’s face.}
\end{figure}

Finally, the scale of a puppet's motion should be only that which is necessary to convey what is intended. Overly precise motions when performing tasks like grasping objects, for example, may risk either attracting too much attention (misconstruing to the audience what is important) or too little attention (making the effort lost on audience members). This is especially true for puppet with many degrees of freedom or complex mechanics. To avoid these risks, puppeteers often 'help' the puppets with tasks without shattering the illusion (Fig.~\ref{fig:details}A). Similarly, gestures that are larger-than-life may be unhelpful unless a puppet is talking passionately. If a puppet uses large gestures when talking about something casually, the audience may feel a disconnect between its body language and attitude  (Fig.~\ref{fig:details}B) .

\subsubsection{Puppeteers' Review}

The professional puppeteers recognize the scale of a puppets movements as being part of an “economy of movement” in which there is a certain amount of movement allowable to convey the emotions of puppet without causing unintended interpretations. However, they also recognized that a puppets morphology may affect the amount of movement that is expected, or needed, in the case of expressive movements.

\begin{quote}
    Puppeteer 2: “I call this, the economy of movement. And keeping things simple. That, like, you want the audience to all understand that it picked up a glass, not that it used its thumb with its fingers to create strength. So yeah, I call this economy of movement. It's really important in clown and magic as well, but just so what people's eyes are drawn to is what we want it drawn to, and that they take away what we're hoping to convey.”
\end{quote}

\begin{quote}
    Puppeteer 3: “Yeah, I agree with both of you guys, because I feel like it is super circumstantial, and a lot of, of styles of puppets are gonna require some over-emoting to get across, what you need, because they don't have all of the features that a human would." 
\end{quote}

\subsubsection{Design Guideline: There is an Economy of Movement for Robots}
Thus far in this paper we have given significant attention to the importance of idle animations. But just as robot designers must consider how to use stillness as a contrast for idle animations, so too must they consider the risks of overly articulated idle animations.
Robot designers should carefully consider what details are necessary --- and unnecessary --- to convey robots' actions, reactions, and emotional states. If a robots' motions overemphasizes details irrelevant to the robot's circumstances, this may lead to confusion and misdirection of attention. For example, articulated humanoid hands may not need to be moving when making large gestures.

\subsubsection{Connections in the Robot Design Literature}
There has been highly relevant recent literature into the behaviors of ``idle hands'' in shopkeeping robots~\cite{pan2024your}, and how to use those hands in a way that is supportive --- however, that work did not explicitly consider the ways that poorly designed idle hand movements might be distracting. Similarly, there is highly relevant research on legibility and predictability from Dragan, which demonstrated that optimal robot motions are not always predictable or legible\cite{dragan2013legibility}. However, this work did not specifically focus on the ways that the degree of variability within the robot's motions may have been distracting. As such, more research is needed in how to encourage perceptions of animacy without unnecessary distraction.

\section{Discussion}
In this work, we leveraged puppetry texts and expert puppetry knowledge to derive a set of eight puppetry principles that can inform robot behavior. The key innovation and contribution of this work is to demonstrate how this overlooked\footnote{Although cp. the recent body of work on this topic, such as that appearing in the 2025 ICRA workshop ``Pulling the Strings on Creative Collaborations: A Retrospective on Puppetry, Choreography, and Control''.} area of artistic practice stands to contribute novel intermediate knowledge for guiding robot behavior design. Specifically, we argue that the intermediate knowledge provided through our analysis serves to fill the gaps between theater-based and dance-based insights. From a top-down perspective, puppeteering insights guide how a robotic character should behave from moment to moment (e.g., breathing, showing emotions, and focusing) within the larger storyboarded interaction scripts that might come out of theater-based design exercises. Dance and movement-based design methods (such as Laban Movement Analysis and Viewpoints theory) and animation-based principles (such as Disney's 12 Principles of Animation) provide concrete heuristics and parameters for designing the animations that visualize these moment-to-moment behaviors. These insights are backed up and validated by the perspectives of the professional puppeteers (most of whom are also actors and improvisers) who served in our focus group.





However, there are some key limitations to this work. First, the texts we leverage focused on live puppetry; insights from filmed puppetry might differ due to the forced framing induced by screens, and the lessened tolerance for visible puppeteers. Moreover, the majority of the texts analyzed focused on approaches like Bunraku in which humanoid puppets are puppeteered; different principles might be followed by those puppeteering smaller puppets, or animal-like puppets.
Finally, in some cases, roboticists may wish to intentionally avoid lifelike behaviors, in order to reinforce robot-likeness. In such cases, some identified principles might need to be modified, minimized, or avoided.

In future work, researchers should explore the use of the eight principles laid out in this work, and identify and quantify the specific benefits brought to the design of different types of robots through their use. Moreover, researchers should explore the ways that these different principles can be leveraged in combination with each other, versus when different principles must be balanced.

Finally, we note that our exploration in this work revealed a number of insights that go beyond robot behavior design, and that relate instead to morphology and interaction design. In future work we hope to explore those further insights in their own right. 

\section{Conclusion}
We have argued that Puppetry represents a fundamentally overlooked artistic practice that has fundamental insights to provide to robot designers, that may complement the types of insights that come from fields like theatre, dance, and animation.
Driven by this insight, we have articulated eight key principles of robot behavior design, confirmed and validated through a focus group with professional puppeteers.

\begin{acks}
    We extend deep thanks to our puppeteer collaborators Thompson Powers, Raquel Hartzell, 
    and Katy Williams.
\end{acks}

\bibliographystyle{ACM-Reference-Format}
\bibliography{references}

\end{document}